\documentclass[11pt]{article}

\usepackage[final]{acl}

\usepackage{microtype}
\usepackage{graphicx}   % 提供 \resizebox
\usepackage{multirow}
\usepackage{booktabs}
\usepackage{adjustbox} 
\usepackage{array}
\usepackage{makecell}
\usepackage{tabularx}
\usepackage{amsmath}
\usepackage{subcaption}

\usepackage{fontspec}
\usepackage{xeCJK}

\newfontfamily\rufont{FreeSerif.ttf}[Path = ./]

\title{All Languages Matter: Understanding and Mitigating \\ Language Bias in Multilingual RAG}

\author{
  \textbf{Dan Wang}\textsuperscript{1,2,$\ast$},
  \textbf{Guozhao Mo}\textsuperscript{1,2,}\thanks{These authors contributed equally.},
  \textbf{Yafei Shi}\textsuperscript{3},
  \textbf{Cheng Zhang}\textsuperscript{3},
  \textbf{Bo Zheng}\textsuperscript{3},
  \textbf{Boxi Cao}\textsuperscript{1,$\dagger$},\\
  \textbf{Xuanang Chen}\textsuperscript{1,}\thanks{Corresponding authors.},
  \textbf{Yaojie Lu}\textsuperscript{1},
  \textbf{Hongyu Lin}\textsuperscript{1},
  \textbf{Ben He}\textsuperscript{1,2},
  \textbf{Xianpei Han}\textsuperscript{1,2},
  \textbf{Le Sun}\textsuperscript{1,2}\\
  \textsuperscript{1}Chinese Information Processing Laboratory,
  Institute of Software, Chinese Academy of Sciences \\
  \textsuperscript{2}University of Chinese Academy of Sciences\\
  \textsuperscript{3}MYbank, AntGroup \\
  \texttt{\{wangdan2023,moguozhao2024,caoboxi,chenxuanang\}@iscas.ac.cn} \\
   \texttt{\{shiyafei.syf,zc481262,guangyuan\}@mybank.cn} \hspace{0.5em} \texttt{benhe@ucas.ac.cn} \\
  \texttt{\{luyaojie,hongyu,sunle,xianpei\}@iscas.ac.cn} 
}

\makeatletter
\renewcommand{\titlebox}{2.5in}
\makeatother

\begin{document}

\maketitle

% \noindent
% $\dagger$~~These authors contributed equally. \\
% $\ast$~~Corresponding author.

\begin{abstract}
% Retrieval-Augmented Generation (RAG) enhances large language models (LLMs) with external knowledge, but its effectiveness remains fragile in multilingual settings. 
% In this work, we present a systematic analysis of multilingual RAG (mRAG) and uncover a performance gap between standard retrieval–reranking-based evidence selection and an answer-optimized upper bound.
% Our analysis reveals that this gap arises from a distributional mismatch: the language distribution induced by multilingual rerankers deviates substantially from the answer-optimized upper-bound distribution, and reranker relevance scores correlate weakly with downstream answer quality.  
% To address this, we propose a language-agnostic reranker training strategy that directly aligns evidence ranking with generation outcomes. By constructing supervision from documents that empirically enable correct answer generation, our approach guides the reranker to prioritize answer-critical evidence over surface-level semantic or linguistic similarity.
% Experiments on MKQA datasets demonstrate that our method consistently reduces language bias and improves mRAG performance across diverse languages and generation models.

% Multilingual Retrieval-Augmented Generation (mRAG) aims to ground large language models (LLMs) in complementary evidence across languages, reflecting the inherently cross-lingual distribution of real-world knowledge.
Multilingual Retrieval-Augmented Generation (mRAG) leverages cross-lingual evidence to ground Large Language Models (LLMs) in global knowledge.
However, we show that current mRAG systems suffer from a language bias during reranking, systematically favoring English and the query's native language.
% However, we identify a prevalent language bias in current reranking pipelines, which disproportionately favor English and the query's native language.
% To assess the potential impact of this bias, we introduce an estimated oracle evidence analysis framework, which reveals a substantial performance gap between standard reranking pipelines and an achievable upper bound.
By introducing an estimated oracle evidence analysis, we quantify a substantial performance gap between existing rerankers and the achievable upper bound.
Further analysis reveals a critical distributional mismatch: while optimal predictions require evidence scattered across multiple languages, current systems systematically suppress such ``answer-critical'' documents, thereby limiting downstream generation performance.
% Further analysis comparing the language distribution of oracle evidence with that of top-ranked documents produced by existing rerankers shows a clear mismatch: documents required for optimal prediction are distributed across multiple languages rather than concentrated in a single one.
% This misalignment indicates that language-biased reranking systematically suppresses answer-critical multilingual evidence, thereby limiting downstream generation performance.
To bridge this gap, we propose \textit{\textbf{L}anguage-\textbf{A}gnostic \textbf{U}tility-driven \textbf{R}eranker \textbf{A}lignment (LAURA)},
% \textit{LAURA} (\textbf{l}anguage-\textbf{a}gnostic \textbf{u}tility-driven \textbf{r}eranker \textbf{a}lignment), 
which aligns multilingual evidence ranking with downstream generative utility. 
Experiments across diverse languages and generation models show that LAURA effectively mitigates language bias and consistently improves mRAG performance.

\end{abstract}

\section{Introduction}

% 检索增强生成（Retrieval-Augmented Generation, RAG）通过在生成过程中引入外部文档证据，已成为提升大模型事实一致性、知识覆盖与可控性的核心技术。随着大模型应用的全球化，多语言RAG已经成为了一个关键的技术。现实世界中的知识并非以单一语言均匀分布，而是天然呈现出跨语言、非对称与互补的结构：大量区域性事实、文化背景、政策信息与技术细节往往仅在特定语言中被系统性记录；即便针对同一主题，不同语言语料在信息粒度、表述重点与时效性方面也存在显著差异。因此，一个真正有效的多语言 RAG 系统，其目标不应仅是支持多语言输入输出，而应能够根据语义相关性，在不同语言的文档之间进行选择与整合，从而为生成模型提供信息量最大的证据集合。

Retrieval-Augmented Generation (RAG), which incorporates external documentary evidence into the generation process, has emerged as a core technique for improving the factual consistency, knowledge coverage, and controllability of large language models (LLMs)~\citep{Lewis_2020, ram-etal-2023-context}.
% Retrieval-Augmented Generation (RAG) has emerged as a central paradigm for improving the factual consistency, knowledge coverage, and controllability of large language models (LLMs) by grounding generation in external document evidence~\citep{Lewis_2020, ram-etal-2023-context}. 
In these cases, multilingual RAG (mRAG) has become a critical technology to address the needs of a global user base for LLMs~\citep{asai2021one, li-etal-2024-bordirlines}.
In real-world settings, knowledge is not uniformly distributed across languages. Instead, it exhibits inherently cross-lingual and complementary structures. Many region-specific facts, cultural contexts, policy details, and technical knowledge are systematically documented only in particular languages. Therefore, an effective multilingual RAG system should go beyond merely supporting multilingual input and output. Its objective should be to select and integrate documents across various languages, thereby providing the generation model with an evidence set that maximizes informational value.

\begin{figure}
    \centering
    \includegraphics[width=0.9\linewidth]{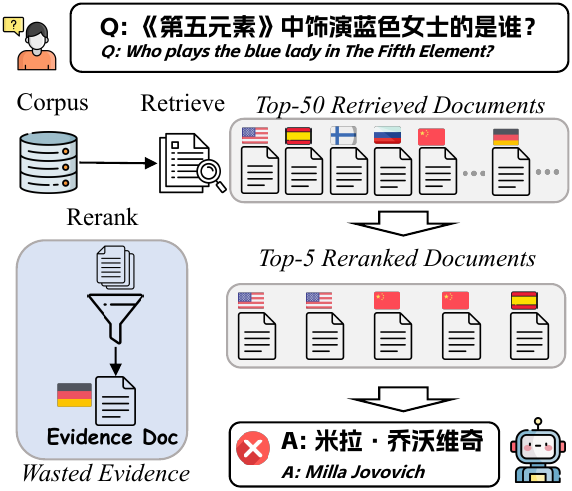}
    \caption{Illustration of failures induced by reranker language bias.}
    \label{fig:compar}
    \vspace{-15pt}
\end{figure}

% 然而，previous studies 已经发现现有的多语言RAG中存在一定的偏差。基于上述的观察，我们对多语言RAG中存在的语言偏好问题进行了系统性的分析，更重要的，与之前不同，我们没有停留在描述这个偏差本身，而是更进一步地探索了这个偏差产生的原因以及其对下游预测的显著影响。
% 有点太口语了

Despite this ideal objective, prior studies have reported the existence of bias in current mRAG systems~\citep{park-lee-2025-investigating, amiraz-etal-2025-cross, qi-etal-2025-consistency}. Motivated by these observations, we present a systematic analysis of language bias in mRAG. Crucially, departing from previous studies that primarily focus on characterizing the presence of bias, we move beyond mere description to investigate the underlying causes of such biases and their significant impact on downstream predictions.

Specifically, based on MKQA dataset, we perform a comprehensive evaluation across multiple rerankers and 13 languages. 
We first construct multilingual candidate document pools and apply standard multilingual retrieval and reranking procedures, after which we analyze the language composition of top-ranked documents. 
Our analysis reveals a consistent pattern: current mRAG systems exhibit a pronounced language preference bias during the reranking stage, systematically favoring English and the original query language. 
For instance, 
when using the widely adopted BGE reranker, more than 70\% of the top-5 retrieved documents, averaged across 13 languages, originate from English and the query language alone. Such a pronounced bias motivated us to dive into its root causes and practical consequences.
 
Conceptually, such language preference bias may stem from two distinct factors. First, it is possible that more accurate or richer information is inherently concentrated in certain languages for specific queries. Second, the bias may arise from the limited multilingual capability of reranking models, which struggle to accurately identify relevant evidence expressed in other languages. Disentangling these two factors is essential for diagnosing the core limitations of current mRAG systems. To this end, we propose a novel multilingual evidence estimation method that approximates the oracle distribution of evidence required to achieve optimal downstream predictions, independent of the reranker’s language preferences.

By comparing estimated oracle evidence distributions, we find that existing multilingual rerankers exhibit limited cross-lingual capability and often fail to provide sufficiently reliable evidence for LLM generation. 
On the MKQA benchmark, standard rerankers underperform the oracle by nearly 20\%, revealing a large performance gap. Further analysis indicates that this gap is not caused by language concentration: oracle evidence is distributed across multiple languages rather than dominated by any single one. Although high-quality evidence already exists in diverse languages within the candidate set, it is systematically downweighted by language-biased rerankers, which substantially limits downstream performance.

To address this misalignment, we propose \textit{\textbf{L}anguage-\textbf{A}gnostic \textbf{U}tility-driven \textbf{R}eranker \textbf{A}lignment (LAURA)}, a training framework that mitigates language bias in multilingual reranking by aligning evidence selection with downstream generation quality. Rather than relying solely on semantic relevance signals, which often favor the query language or high-resource languages, LAURA derives supervision from multilingual documents that lead to better generation outcomes in practice. It then trains the reranker to prioritize answer-critical evidence regardless of language. This utility-driven alignment reduces systematic language preferences in evidence selection and yields consistent improvements in generation performance.

Our major contributions are summarized as follows:
\begin{itemize}
    \item We systematically investigate and quantify language bias in mRAG. We further introduce an estimated oracle evidence analysis framework, revealing that such bias substantially constrains the generation performance of mRAG systems.
    \item We propose LAURA, an answer-utility-driven reranking framework that leverages generation outcomes as supervision signals. LAURA effectively mitigates language bias while consistently improving downstream task performance.
\end{itemize}

\section{Reranking Bias in mRAG Systems}

\label{sec:limit-of-mrag}

\begin{figure*}[t]
  \centering
  \includegraphics[width=\textwidth]{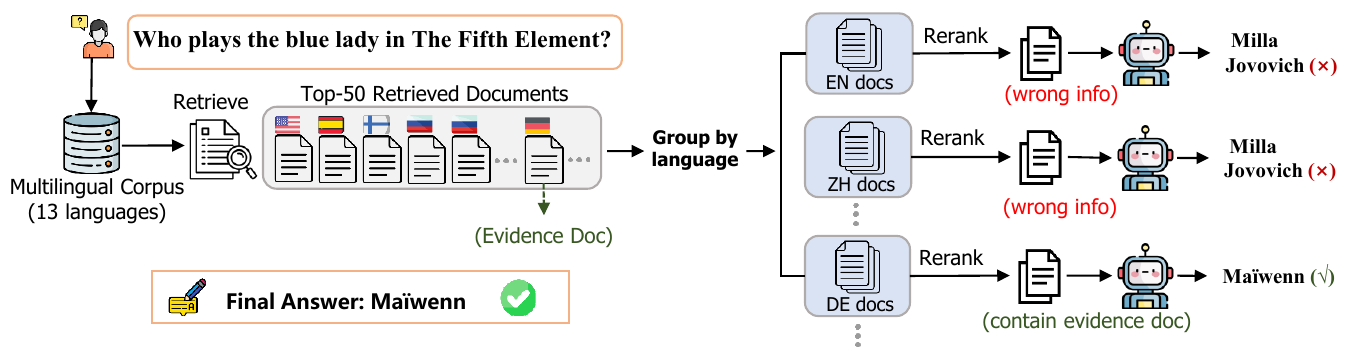}
  \caption{Illustration of the oracle evidence estimation strategy, where candidate documnents are grouped by language and reranked independently to select the top-5 documents within each language group, and multilingual evidence documents are selected based on correctness of the generated answer.
  }
  \label{fig:main-fig}
  \vspace{-10pt}
\end{figure*}

While prior work has identified performance degradation in mRAG, it largely focuses on pipeline-level optimizations, such as translation-based strategies, without rigorously quantifying the system’s theoretical upper bound or identifying the underlying causes. A key unresolved question is whether current bottlenecks arise from insufficient relevant information in the retrieval pool or from the selection mechanism’s inability to identify accurate multilingual evidence. To bridge this gap, we present a systematic analysis comparing standard retrieval pipelines against an oracle evidence estimating setting, aiming to reveal the misalignment between relevance-based selection and actual answer utility.

\subsection{Language Distribution Analysis}
\label{sec:Experimental-Setups}

To quantify the limitations of current mRAG pipelines, we define two contrasting settings and a method for analyzing language distribution.

\paragraph{Vanilla Document Reranking.}
Following the standard multilingual RAG setup adopted in previous work~\citep{chirkova-etal-2024-retrieval}, 
for each query $q \in \mathcal{Q}$, we retrieve documents from a unified multilingual corpus that contains documents from all evaluation languages (13 languages in total).
The pipeline consists of two stages: first, a multilingual retriever BGE-M3~\citep{Chen_Xiao_Zhang_Luo_Lian_Liu} fetches the top-50 candidate passages across all languages; second, a multilingual reranker, such as BGE-Reranker-V2-M3~\citep{Chen_Xiao_Zhang_Luo_Lian_Liu} and Qwen3-Reranker-0.6B~\citep{qwen3embedding}, selects the top-5 most relevant passages. These passages are concatenated to form the context for the generator. The quality of the generated answers is evaluated using the metrics defined below.

\paragraph{Oracle Evidence Estimating.}
As show in Figure~\ref{fig:main-fig}, to estimate the performance upper bound given the retrieved candidates, we adopt a language-wise reranking strategy.
For a query $q \in \mathcal{Q}$, the pool of 50 retrieved candidates is partitioned by document language.
Within each language group, we select the top-5 documents (or fewer if insufficient candidates exist) to generate a language-specific answer.
The final performance for query $q$ is defined as the maximum score achieved across all language groups, serving as an estimated upper limit for language selection. We use BGE-M3 embeddings for retrieval and the BGE-Reranker-V2-M3 for reranking.

\paragraph{Language Distribution Computation.} To understand the linguistic composition of selected evidence, we calculate distribution metrics for both settings:
\begin{itemize}
  \item \textbf{Vanilla Distribution.}
  For each query, we calculate the proportion of each language within the final top-5 documents chosen by the reranker (e.g., three English and two Chinese documents yield a distribution of 0.6 and 0.4, respectively).
  These per-query distributions are then averaged over all queries in a specific query language to obtain the overall context language distribution.

  \item \textbf{Oracle Distribution.}
  For each query, we identify the document language(s) that produce the best-performing answer.
  We assign an importance weight to languages based on answer performance: if a single language achieves the best score, it receives a weight of 1; if multiple languages tie for the best, the weight is uniformly distributed among them (e.g., a tie between English and Chinese results in 0.5 for each).
  Similar to the vanilla setting, these per-query weights are averaged across all queries for each query language.
\end{itemize}

\subsection{Experimental Setups}
\paragraph{Datasets.}
For the multilingual document corpus, we use English Wikipedia\footnote{\url{https://huggingface.co/datasets/facebook/kilt_wikipedia}} and Wikipedia in the corresponding user languages\footnote{\url{https://huggingface.co/datasets/wikimedia/wikipedia}}.
Following the preprocessing strategy of~\citep{chirkova-etal-2024-retrieval}, we split each Wikipedia article into chunks of 100 words.
For languages without explicit whitespace segmentation, namely Chinese, Japanese, and Thai, we instead split articles into chunks of 100 Unicode characters.
The article title is prepended to each chunk.

\begin{figure*}[t!]
  \centering
  \includegraphics[width=\textwidth]{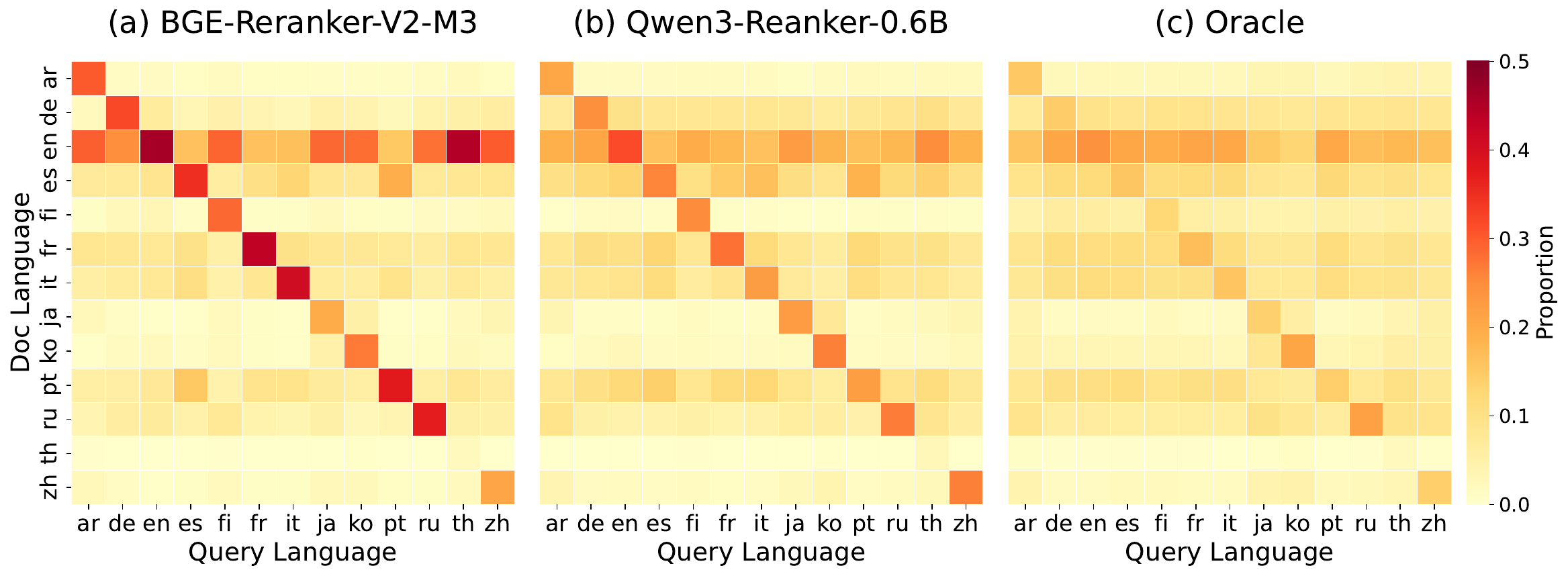}
  \caption{Heatmaps showing the proportion of selected document languages (y-axis) for each query language (x-axis). (a): Distribution from the BGE-Reranker-V2-M3 reranker. (b): Distribution from the Qwen3-Reranker-0.6B reranker. (c): The oracle evidence distribution derived from our estimation strategy. Results for other reranking models are detailed in Appendix~\ref{sec: all-lang-dist-fig}.
  }
  \label{fig:qwen-baseline-upper}
  \vspace{-10pt}
\end{figure*}

For multilingual question answering, we use the MKQA~\citep{Longpre_Lu_Daiber_2021} dataset, following the setup of~\citet{chirkova-etal-2024-retrieval}.
MKQA is a multilingual open-domain QA benchmark consisting of 10,000 questions from the Natural Questions (NQ) dataset~\citep{Kwiatkowski_NQ}, translated into 25 languages.
% The dataset is parallel across languages and primarily grounded in English Wikipedia.
In our experiments, we focus on a subset of languages used for evaluation.
Specifically, we select 2.7K samples that overlap between MKQA and the KILT NQ dataset\footnote{\url{https://huggingface.co/datasets/facebook/kilt_tasks}}, enabling access to corresponding document-level relevance information for the selected test languages. 
% XOR-TyDi QA is a cross-lingual open-domain QA benchmark comprising approximately 40K information-seeking questions in 7 languages. Each question is grounded in Wikipedia either in the same language as the query or in English. Following prior work, we evaluate on the 3K-question validation set. To provide English for comparison, we include results for English on the TyDi QA dataset~\citep{Clark_Choi_Collins_Garrette_Kwiatkowski_Nikolaev_Palomaki_2020}.

\paragraph{Models.}
For retrieval, we use BGE-M3~\citep{Chen_Xiao_Zhang_Luo_Lian_Liu}, a strong and publicly available multilingual embedding model capable of encoding all languages considered in our experiments.

For reranking, we adopt BGE-Reranker-V2-M3 and Qwen3-Reranker-0.6B~\citep{qwen3embedding} as representatives of mainstream encoder-only rerankers and LLM-based rerankers, respectively.

For answer generation, we evaluate two multilingual large language models, including Qwen2.5-7B-Instruct~\citep{qwen2025qwen25technicalreport}, and Llama-3.1-8B-Instruct~\citep{grattafiori2024llama3herdmodels}.

\paragraph{Evaluation Metric.}
Following~\citet{chirkova-etal-2024-retrieval}, we evaluate model outputs using the character-level 3-gram recall metric. The details are shown in Appendix ~\ref{sec:metric_details}.

\subsection{Analysis Results}

\subsubsection{Multilingual Rerankers Exhibit Systemic Language Bias}
\paragraph{Conclusion 1.} \textit{Current multilingual RAG systems exhibit a pronounced language preference bias during the reranking stage, systematically favoring English and the original query language.}

To understand the linguistic preferences of current mRAG systems, we analyze the language distribution of the documents selected for generation. As illustrated in Figure~\ref{fig:qwen-baseline-upper}, the heatmaps display two dominant patterns: a strong diagonal alignment reflecting a bias toward the query language, and a pronounced horizontal alignment indicating a systemic preference for English. Taking BGE-Reranker as a example, around 60\% of candidate documents are concentrated in English and the query language. This distribution confirms that current rerankers heavily prioritize documents based on surface-level language matching or dominant language priors (predominantly English), rather than assessing semantic relevance equitably across all candidate languages.

\subsubsection{Reranking Bias as a Primary Performance Bottleneck}
\paragraph{Conclusion 2.} \textit{These reranking biases constitute a primary performance bottleneck in multilingual RAG by causing the model to overlook genuinely relevant evidence within the candidate pool, thereby hindering the retrieval of optimal information.}

To determine whether this pronounced language bias stems from an intrinsic concentration of high-quality information in dominant languages or a fundamental lack of multilingual capability in current rerankers, we conduct a decoupled analysis by contrasting the standard pipeline with an \textit{Oracle Evidence Estimating} setting. This comparison allows us to isolate the model's selection bias from the quality of the candidate pool, thereby identifying the core defect in the current evidence selection mechanism.

First, to quantify the extent to which reranking limits system performance, we evaluated the generation quality under both settings and computed the correlation between reranking scores and answer utility. As shown in Table~\ref{tab:baseline_upper_limit}, simply selecting the correct documents from the existing retrieval pool yields substantial improvements ranging from +12.9 to +20 points. This result confirms that the retrieval stage successfully recalls the necessary information, but the reranker fails to surface it. Furthermore, quantitative analysis reveals a weak correlation between reranker relevance scores and downstream answer quality, with Pearson coefficients consistently below 0.2 across all models (Table~\ref{tab:rerank_corr}). 
\textbf{These indicate that current multilingual rerankers fail to provide sufficiently accurate and effective evidence, creating a bottleneck that strictly limits the generation potential of LLMs.}

Next, to understand why valid evidence is overlooked, we analyzed the language distribution of estimated oracle evidence and conducted a case study to observe model behavior. By analyzing the language distribution under the Oracle Evidence Estimating setting (Figure~\ref{fig:qwen-baseline-upper}, c), we find that true answer-critical evidence is broadly distributed across diverse, non-query languages, rather than being concentrated in the query language. However, the systemic bias identified in the previous section filters these optimal documents out. This phenomenon is exemplified in the Case Study (Table~\ref{tab:case_study}): for the query "Who plays the blue lady in The Fifth Element?", the reranker prioritizes non-informative query-language documents (ranks 1-5) leading to hallucination, while suppressing the decisive multilingual evidence to rank 10. 
\textbf{Thus, while genuinely relevant evidence is already present within candidate documents across diverse languages, it is consistently marginalized by the systemic language preferences of current rerankers, thereby significantly constraining downstream performance.}

\begin{table}[t]
\centering
\small
\resizebox{0.48\textwidth}{!}{
\begin{tabular}{l|ccc|ccc}
\toprule
\multirow{2}{*}[-0.8ex]{\textbf{Lang}} & \multicolumn{3}{c|}{\textbf{Llama-8B-Instruct}} 
& \multicolumn{3}{c}{\textbf{Qwen2.5-7B-Instruct}} \\
\cmidrule(lr){2-4} \cmidrule(lr){5-7}
& \textbf{BGE} & \textbf{Qwen3} & \textbf{Oracle} 
& \textbf{BGE} & \textbf{Qwen3} & \textbf{Oracle} \\
\midrule
ar & 32.7 & 28.9 & 53.6 & 33.8 & 31.4 & 51.4 \\
de & 62.8 & 60.9 & 76.5 & 59.6 & 58.0 & 73.7 \\
en & 70.1 & 67.8 & 79.3 & 65.4 & 63.3 & 76.2 \\
es & 63.0 & 62.7 & 76.8 & 62.3 & 61.3 & 75.6 \\
fi & 58.1 & 54.2 & 73.4 & 55.5 & 52.8 & 71.5 \\
fr & 64.4 & 63.7 & 76.7 & 56.6 & 54.4 & 71.6 \\
it & 63.9 & 62.2 & 77.0 & 60.2 & 57.9 & 73.8 \\
ja & 29.2 & 28.2 & 47.9 & 28.0 & 27.2 & 44.9 \\
ko & 25.5 & 23.3 & 41.0 & 26.5 & 24.7 & 38.9 \\
pt & 66.4 & 66.3 & 78.4 & 60.5 & 59.3 & 73.8 \\
ru & 51.9 & 47.4 & 68.0 & 45.7 & 42.3 & 63.1 \\
th & 26.4 & 24.8 & 44.1 & 23.5 & 22.3 & 39.0 \\
zh & 21.7 & 21.9 & 33.8 & 29.0 & 28.7 & 42.8 \\
\midrule
\textbf{AVG} & 48.9 & 47.1 & 63.6 & 46.7 & 44.9 & 61.3 \\
\bottomrule
\end{tabular}
}
\caption{Performance comparison (Recall@3-gram) of vanilla reranking and oracle evidence estimating. `BGE' and `Qwen3' refer to BGE-Reranker-V2-M3 and Qwen3-Reranker-0.6B models, respectively. `Oracle' denotes the performance achieved under the estimated oracle evidence.}
\label{tab:baseline_upper_limit}
\vspace{0pt}
\end{table}

\begin{table}[t]
\centering
\small
\resizebox{0.48\textwidth}{!}{
% \begin{tabular}{l l c c}
\begin{tabular}{llcc}
\toprule
\textbf{Reranker} & \textbf{Model} & \textbf{Pearson} & \textbf{p-value} \\
\midrule
\multirow{2}{*}{BGE-Reranker}
 & Llama3-8B-Instruct & 0.188 & $3.8 \times 10^{-290}$ \\
 & Qwen2.5-7B-Instruct & 0.198  & $1.0 \times 10^{-320}$ \\
\midrule
\multirow{2}{*}{Qwen-Reranker}
 & Llama3-8B-Instruct & 0.129 & $1.2 \times 10^{-135}$ \\
 & Qwen2.5-7B-Instruct & 0.127 & $2.3 \times 10^{-135}$ \\
\bottomrule
\end{tabular}
}
\caption{Correlation between relevance scores (mean top-5) and downstream answer performance (Recall@3-gram) under different rerankers and generators.}
\label{tab:rerank_corr}
\vspace{-15pt}
\end{table}

\section{Language-Agnostic Utility-driven Reranker Alignment}
In this section, we aim to mitigate language bias in multilingual rerankers. Such bias leads models to disproportionately favor documents in English or the query language, even when higher quality evidence exists in other languages. We hypothesize that skewed training data, in which high quality query and document annotations are scarce for low resource languages, is a key factor driving this disparity.

\begin{figure*}
    \centering
    \includegraphics[width=0.8\linewidth]{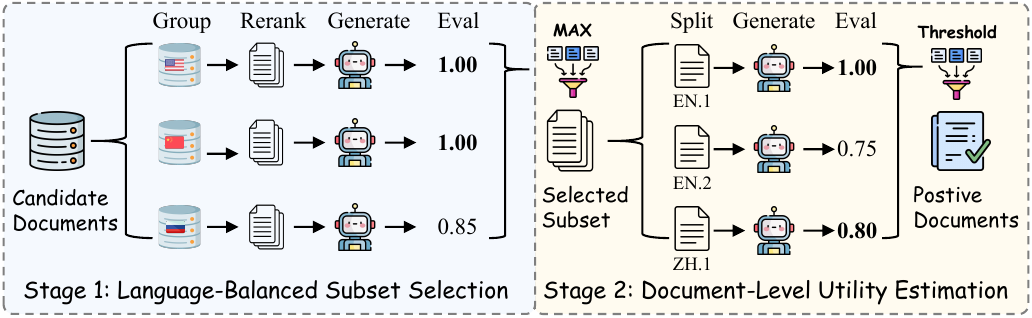}
    \vspace{-5pt}
    \caption{Two-stage data construction pipeline in the LAURA framework.}
    \vspace{-5pt}
    \label{fig:data_pipeline}
\end{figure*}

To address this issue, we propose a \textit{language agnostic utility driven reranker alignment} framework (LAURA). This framework reduces language bias by grounding reranker supervision in answer utility instead of relying on language dependent relevance signals. Instead of defining positives based on lexical overlap or language matching, LAURA selects documents according to their contribution to downstream answer quality, thereby reducing reliance on language specific surface features. Specifically, LAURA uses a two stage data construction pipeline (Figure~\ref{fig:data_pipeline}) to generate language agnostic supervision signals, followed by listwise reranker fine tuning. This design promotes balanced cross lingual supervision and aligns reranker preferences with answer correctness, thereby mitigating the over preference for high resource languages.

\subsection{Answer Utility-driven Data Construction}
 % for Language Bias Mitigation
Although many RAG QA datasets are publicly available, most only provide annotations for the correctness of the final answer, without explicit query-document relevance labels. This absence leads to language bias in reranker training. We aim to automatically generate such annotations while maintaining balanced multilingual coverage.

Given a query $q$, we retrieve a candidate document set $D$ from a multilingual corpus. Our objective is to select a positive subset $D_{\text{pos}} \subset D$ consisting of documents that genuinely support answering the query, free from language-specific bias. We use the average answer quality produced by multiple generators conditioned on a document as a proxy for its answer utility.

\paragraph{Stage 1: Language-Debiased Subset Selection.}
Directly estimating answer utility on top-ranked retrieved documents can amplify the inherent language bias of multilingual rerankers, which often favor documents in high-resource or query-matched languages. To mitigate this effect, we propose a candidate debiasing stage that filters the candidate set before utility estimation while preserving overall utility.

Given a retrieved document set $D$, we partition documents into disjoint subsets according to their language and apply the same reranker to rank documents \emph{within each subset independently}. From each subset, we retain up to five top-ranked documents as utility candidates. This procedure does not assume language-specific relevance. Instead, it enforces equal exposure across linguistic subsets, preventing the candidate pool from being dominated by documents favored due to language priors rather than informational content.

The retained documents are then evaluated by multiple generators to estimate their average answer quality. Documents achieving the highest generation utility are selected for subsequent supervision construction. In cases where multiple subsets yield identical maximal utility (e.g., all generators produce correct answers), all corresponding documents are preserved. The resulting candidate set is denoted as $D_{\text{balanced}}$.

By decoupling candidate selection from global reranker scores and restricting comparisons to within each subset, this stage reduces language-induced ranking bias while retaining documents that are useful for downstream answer generation.

\paragraph{Stage 2: Document-Level Utility Estimation.}
While Stage~1 ensures cross-lingual coverage, documents in $D_{\text{balanced}}$ may still vary in their actual usefulness. We therefore perform fine-grained document-level utility estimation by evaluating each document independently via generation.

To avoid introducing an implicit language bias through relative ranking alone, we apply an absolute utility threshold $\theta$ and retain only documents whose average generation performance exceeds this threshold. The final positive set $D_{\text{pos}}$ thus consists of documents that demonstrably contribute to answer correctness, independent of language.

Overall, this two-stage procedure yields high-quality, language-debiased training data that grounds reranker supervision in answer utility rather than language preference.

\subsection{Listwise Reranker Fine-Tuning}

Using the constructed training data, we fine-tune the reranker with a listwise learning objective. Given a query $q$ and a candidate set $D$, documents not selected into $D_{\text{pos}}$ are treated as negatives, forming $D_{\text{neg}}$. 

During training, we construct training instances consisting of one positive document and $k$ negative documents, i.e., $(q, d_{\text{pos}}, \{d_{\text{neg}}^{(i)}\}_{i=1}^{k})$, where $d_{\text{pos}} \in D_{\text{pos}}$ and $d_{\text{neg}}^{(i)} \in D_{\text{neg}}$. The reranker produces a relevance score $s(q, d)$ for each document $d \in \mathcal{D}_q$. Training encourages the model to assign the highest score to the positive document within the list. We adopt a softmax cross-entropy loss:
\begin{equation}
\mathcal{L}
= - s(q, d_{\text{pos}})
+ \log \sum_{d \in \mathcal{D}_q} \exp \bigl(s(q, d)\bigr)
\end{equation}

For encoder-only rerankers, $s(q, d)$ is produced directly as a scalar logit. For LLM-based rerankers, the score is derived from the relative logits of predefined
\texttt{positive\_token} and \texttt{negative\_token}, which represent the model’s
preference over relevance labels.
\subsection{Experimental Setups}
\paragraph{LAURA Dataset.} We use data from the MKQA benchmark, selecting only samples that are disjoint from the evaluation test set to avoid data leakage. 

In stage 1, for each question, we retrieve the top-100 candidate documents from a multilingual Wikipedia corpus using the BGE-M3 retriever. 
Within each language group, we apply the multilingual BGE reranker to select the top-5 documents, yielding a language-debiased candidate set. 

During evaluation, we prompt multiple generation models to answer the question conditioned on each document independently and measure answer quality using character-level 3-gram recall. To reduce model-specific bias, we compute each document’s utility score as the average generation performance across a diverse set of four generation models, including Qwen2.5-7B-Instruct, Qwen2.5-14B-Instruct, Llama3-8B-Instruct, and DeepSeek-R1-Distill-Qwen-7B~\citep{deepseekai2025deepseekr1incentivizingreasoningcapability}. The threshold $\theta$ is set to 0.8, ensuring that the documents retain high utility. 

Finally, we construct a total of \textbf{18,360} query--positive documents pairs. Among them, \textbf{1,000} are randomly sampled as the dev set. Detailed statistics of the constructed fine-tuning dataset are reported in Appendix~\ref{sec:stat-of-finetuning-dataset}.

\paragraph{Evaluation Metric.}
To evaluate the effectiveness of LAURA, we adopt Precision@k and NDCG@k to assess the rerank performance on positive documents in the dev set. In addition, we use the PEER~\citep{peer} metric to measure whether the reranker exhibits language-specific bias. PEER is based on the assumption that documents with equal relevance should have similar average rankings across different languages. Higher PEER scores indicate weaker language preference. The detailed definitions of the evaluation metrics are provided in Appendix~\ref{sec:metric_details}.

\paragraph{Training Details.}
We fine-tune BGE-Reranker-V2-M3 using the implementation provided by FlagEmbedding\footnote{\url{https://github.com/FlagOpen/FlagEmbedding}}, and Qwen3-Reranker-0.6B using SWIFT~\citep{swift}. 
For each query, BGE is trained with 1 negative document,
whereas Qwen uses 7 negative documents,
reflecting the stronger capacity of the LLM-based reranker to handle larger candidate lists.
Both models are optimized with AdamW~\citep{adamw}, using a learning rate of $6\times10^{-6}$, and are trained for five epochs.

\subsection{Results of LAURA}

\begin{table}
    \centering
    \resizebox{0.48\textwidth}{!}{
    \begin{tabular}{lccccc}
        \toprule
        \multirow{2}{*}[-0.8ex]{\textbf{Setting}} & \multicolumn{2}{c}{\textbf{Precision}} & \multicolumn{2}{c}{\textbf{NDCG}}& \multirow{2}{*}[-0.8ex]{\textbf{PEER}}  \\
\cmidrule(lr){2-3} \cmidrule(lr){4-5}
         & \textbf{@5} & \textbf{@10} & \textbf{@5} & \textbf{@10} &  \\ 
        \midrule
        BGE-Reranker & 0.3400 & 0.2712 & 0.4666 & 0.4904& 0.5941 \\ 
        \quad\textbf{+ LAURA}  & \textbf{0.3830} & \textbf{0.3149} & \textbf{0.5531} & \textbf{0.5925}&\textbf{0.6627} \\ 
        \midrule
        Qwen-Reranker & 0.2702 & 0.2206 & 0.3695 & 0.3921&0.6606 \\ 
        \quad\textbf{+ LAURA}  & \textbf{0.3546} & \textbf{0.2847} & \textbf{0.5214} & \textbf{0.5496}&\textbf{0.6720} \\ 
        \bottomrule
    \end{tabular}}
    \caption{Reranking results of BGE-Reranker-V2-M3 (BGE-Reranker) and Qwen3-Reranker-0.6B (Qwen-Reranker) on the dev set before and after LAURA training. PEER measures language bias in the reranker, with higher values indicating weaker language preference.}
    \vspace{-15pt}
\label{tab:dev_result}
\end{table}

\paragraph{LAURA improves multilingual rerankers' ability to identify relevant documents.}
To evaluate whether rerankers can better identify positive candidates under the LAURA, we assess Precision, NDCG on the dev set before and after training. These metrics directly reflect the rerankers' ability to rank relevant candidates higher.
As shown in Table~\ref{tab:dev_result}, both BGE and Qwen rerankers exhibit consistent improvements after being trained within the LAURA. In particular, Precision@5 increases by approximately 6 points, while NDCG@5 improves by around 13 points across both model families, indicating a stronger capability to place positive candidates at higher ranks.

\paragraph{LAURA improves multilingual rerankers' language fairness.}
We observe that LAURA leads to consistent improvements in language fairness. Beyond the quantitative gains on the dev set measured by the PEER metric, we analyze the language distribution of reranker outputs on the MKQA test set after LAURA training. As shown in Table~\ref{tab:distance_table}, the JS divergence and KL divergence between the post-training distribution and the estimated oracle evidence distribution are substantially reduced, demonstrating that the learned distribution moves closer to the desired target distribution. Moreover, we observe a consistent decrease in the proportion of documents written in English and the query language, suggesting that LAURA mitigates the over-preference for dominant languages and encourages a more balanced multilingual ranking behavior. This indicates that LAURA effectively reduces the original language skew of rerankers. In terms of PEER, LAURA yields an about +7 points for the BGE reranker and about +0.5 points for the Qwen reranker, suggesting that the method systematically mitigates language biases and promotes more equitable performance across languages.

\begin{table}
    \centering
    \resizebox{0.4\textwidth}{!}{
    \begin{tabular}{lccc}
    \toprule
    \textbf{Setting}& \textbf{JS} & \textbf{KL} &\textbf{Entropy}  \\
    \midrule
      BGE-Reranker &0.203&0.186& 2.03 \\ 
    \quad\textbf{+ LAURA}  &\textbf{0.090}&\textbf{0.041} & \textbf{2.27} \\ 
    \midrule
    Qwen-Reranker &0.141&0.122&2.13  \\ 
    \quad\textbf{+ LAURA}  &\textbf{0.129}&\textbf{0.094} & \textbf{2.14}  \\ 
    \bottomrule
    \end{tabular}
    }
    \caption{Language distributional metrics before and after LAURA training on the MKQA test set. JS and KL denote the average distances between vanilla distribution and the estimated oracle distribution. Entropy indicates the average entropy of the vanilla distribution of each query language.}
    \vspace{-10pt}
    \label{tab:distance_table}
\end{table}

\paragraph{LAURA improves downstream generation performance and ranking utility.}
LAURA is designed to enhance reranking quality and improve the alignment between reranking scores and downstream generation. To investigate to what extent the improved reranking capability learned under the LAURA transfers to downstream generation performance, we conduct experiments on the MKQA test set using the setup in Section~\ref{sec:Experimental-Setups}. The results are reported in Table~\ref{tab:qa_res_train}. On the 3-gram recall metric, incorporating LAURA leads to an average improvement of 1.95 points for the Qwen reranker and 1.0 points for the BGE reranker. These results indicate that improving the rerankers' ability to select higher-quality candidates can translate into better downstream generate quality.

To quantitatively assess the change in the relationship between ranking quality and generation performance, we compute the Pearson correlation between the average reranking score of the top-5 documents and the corresponding 3-gram recall scores. After training with LAURA, the Pearson correlation increases by approximately 25\% for the BGE reranker and by about 108\% for the Qwen reranker. This demonstrates that LAURA substantially strengthens the correlation between reranking scores and generation performance, thereby improving the practical utility of the reranking scores for downstream generation.

\begin{table}[t]
\centering
\small
\resizebox{0.48\textwidth}{!}{
\begin{tabular}{lcccc}
\toprule
\multirow{2}{*}[-0.8ex]{\textbf{Setting}} & \multicolumn{2}{c}{\textbf{Llama}} & \multicolumn{2}{c}{\textbf{Qwen}} \\
\cmidrule(lr){2-3} \cmidrule(lr){4-5}
 & \textbf{3-gram} & \textbf{Pearson}& \textbf{3-gram} & \textbf{Pearson} \\
\midrule
    BGE-Reranker  &  48.9 &0.198 & 46.7 &0.188 \\
    \quad\textbf{+ LAURA} & \textbf{49.9} &\textbf{0.236} & \textbf{47.7} & \textbf{0.247}\\ 
    \midrule
    Qwen-Reranker  & 47.1&0.129 & 44.9&0.127 \\ 
    \quad\textbf{+ LAURA} & \textbf{49.2}& \textbf{0.269} & \textbf{46.7}& \textbf{0.264}\\ 
\bottomrule
\end{tabular}
}
\vspace{0pt}
\caption{Generation performance and Pearson correlation of rerankers before and after LAURA training on the MKQA test set. Pearson correlations are computed between the average reranker scores of the top-5 reranked documents and character 3-gram recall performance. All Pearson correlations are statistically significant with p-values < 0.001.}
\label{tab:qa_res_train}
\end{table}

\begin{table}
    \centering
    \small
    \resizebox{0.48\textwidth}{!}{
    \begin{tabular}{lcccc}
    \toprule
    \multirow{2}{*}[-0.8ex]{\textbf{Setting}} & \multicolumn{2}{c}{\textbf{Llama}} & \multicolumn{2}{c}{\textbf{Qwen}} \\
    \cmidrule(lr){2-3} \cmidrule(lr){4-5}
     & \textbf{3-gram} & \textbf{Pearson}& \textbf{3-gram} & \textbf{Pearson} \\
    \midrule
        BGE-Reranker  &  48.9 &0.198 & 46.7 &0.188 \\
        Self-Training  &  48.9 &0.188 & 46.7 &0.202 \\
        mMARCO & 48.7 & 0.132 & 46.3 & 0.137 \\
        LAURA & \textbf{49.9} &\textbf{0.236} & \textbf{47.7} & \textbf{0.247}\\ 
    \bottomrule
    \end{tabular}
    }
    \caption{Performance comparison of LAURA against alternative fine-tuning strategies, including Self-Training (naive supervision using top-5 retrieved candidates) and mMARCO fine-tuning (general-purpose multilingual ranking data).}
    \label{tab:blank_baseline}
\end{table}

\subsection{Comparison Against Fine-tuning Baselines}

We provide additional analysis on two alternative fine-tuning strategies to further validate the effectiveness of LAURA's data construction pipeline.

\paragraph{Self-Training Baseline.}

The first baseline fine-tunes the reranker solely on its own top-ranked outputs as pseudo-positive supervision, directly treating the top-5 re-ranked documents as relevant and all remaining candidates as non-relevant, without any additional filtering or refinement. This setting corresponds to the starting point of LAURA's data construction pipeline and serves as an \textit{empty control} to verify whether LAURA's additional filtering and refinement steps contribute beyond naive supervision. Under this paradigm, the model's existing ranking preferences may be progressively reinforced, as no mechanism is introduced to correct noisy or biased pseudo labels.

\paragraph{Fine-tuning on mMARCO.}

The second baseline fine-tunes the reranker using mMARCO \cite{mmarco}, a widely-used multilingual dataset, to examine whether general-purpose training data can address the specific distribution imbalance in mRAG. We randomly sample 20k queries from mMARCO for training, comparable to the 17,360 queries used by LAURA, ensuring a fair comparison in terms of training scale.

For both baselines, we use the same hyperparameters as in our main experiments. In addition, we ensure that LAURA and the Self-Training baseline are trained on the exact same set of queries, isolating the effect of the data construction strategy rather than the training queries themselves.

As shown in Table~\ref{tab:blank_baseline}, LAURA consistently outperforms both baselines across all settings. The Self-Training baseline fails to surpass BGE-Reranker on certain metrics, indicating that naive pseudo-label supervision can reinforce existing biases rather than correct them. The mMARCO baseline also leads to a slight performance drop compared to BGE-Reranker, suggesting that general relevance signals cannot resolve the specific distribution imbalance in mRAG. These results collectively demonstrate that LAURA's filtering and refinement steps are essential for effective reranker adaptation in mRAG settings.

\section{Related Work}
mRAG is pivotal for bridging global information gaps and ensuring equitable knowledge access across linguistic barriers. To advance this capability, the community has established a robust foundation spanning diverse benchmarks~\citep{asai-etal-2021-xor, 10.5555/3540261.3540839, liu-etal-2025-xrag} and retrieval architectures~\citep{gao-etal-2022-retrieval, 10.1145/3613447, chirkova-etal-2024-retrieval}.

% \noindent \textbf{Bias Indentification}
Previous studies have conducted preliminary analyses of language preference phenomena in mRAG systems. For instance, \citet{amiraz-etal-2025-cross} investigates multilingual retrieval biases over Arabic–English corpora. \citet{park-lee-2025-investigating} evaluate language bias in multilingual RAG by measuring retrieval ranking shifts.
% Recent studies have identified systematic language biases in these systems, observing a tendency for retrievers and rerankers to favor documents in the query language or high-resource languages like English~\citep{park-lee-2025-investigating, amiraz-etal-2025-cross, qi-etal-2025-consistency}. 
% These biases often limit the utilization of informative evidence from low-resource or linguistically distant languages~\citep{wu2024limitscrosslingualdensepassage, sharma-etal-2025-faux}. 
In comparison, our work moves beyond merely characterizing these preferences to systematically quantify the substantial performance gap resulting from this linguistic misalignment.

To mitigate these biases, prior research has largely relied on translation-centric strategies, such as mapping queries or documents to a shared pivot language~\citep{moon-etal-2025-quality, amiraz-etal-2025-cross, park-lee-2025-investigating}. However, these pipeline-level heuristics depend heavily on the capability of translation models and do not fundamentally correct the ranking objective. 
In comparison, we propose to align rerankers directly with generation utility, training the model to prioritize answer-critical evidence regardless of the source language.
\section{Conclusion}
This work analyzes language bias in multilingual retrieval-augmented generation (mRAG) systems, showing that conventional rerankers favor English and the query’s original language, suppressing critical multilingual evidence. Using estimated oracle evidence, we reveal the resulting performance gap and cross-lingual distribution of answer-relevant documents. To address this, we propose LAURA, a language-agnostic utility-driven reranker that aligns evidence ranking with downstream generation, mitigating bias and improving performance across languages and models.

\section*{Limitations}
This work focuses on analyzing the alignment between reranker relevance and downstream answer quality in multilingual RAG systems. Accordingly, our study is limited to the reranking stage and does not consider modifications to the retriever or the generator, whose interactions with reranking remain an important direction for future work.

In addition, our evaluation relies on automatic, task-specific metrics that may not fully capture all aspects of generation utility, such as factual completeness or cross-lingual reasoning. Finally, while our experiments cover diverse multilingual settings, the generalizability of our findings to other architectures, domains, and low-resource languages warrants further investigation.

\section*{Acknowledgments}
We sincerely thank the reviewers for their insightful comments and valuable suggestions. This work was supported by Beijing Natural Science Foundation (L243006), the Natural Science Foundation of China (No. 62536008, 62506354), the Postdoctoral Fellowship Program of CPSF under Grant Number GZC20251041, and MYbank, AntGroup.

\bibliography{main}

\begin{thebibliography}{25}
\providecommand{\natexlab}[1]{#1}

\bibitem[{Amiraz et~al.(2025)Amiraz, Fyodorov, Haramaty, Karnin, and Lewin-Eytan}]{amiraz-etal-2025-cross}
Chen Amiraz, Yaroslav Fyodorov, Elad Haramaty, Zohar Karnin, and Liane Lewin-Eytan. 2025.
\newblock \href {https://doi.org/10.18653/v1/2025.arabicnlp-main.6} {The cross-lingual cost: Retrieval biases in {RAG} over {A}rabic-{E}nglish corpora}.
\newblock In \emph{Proceedings of The Third Arabic Natural Language Processing Conference}, pages 69--83, Suzhou, China. Association for Computational Linguistics.

\bibitem[{Asai et~al.(2021{\natexlab{a}})Asai, Kasai, Clark, Lee, Choi, and Hajishirzi}]{asai-etal-2021-xor}
Akari Asai, Jungo Kasai, Jonathan Clark, Kenton Lee, Eunsol Choi, and Hannaneh Hajishirzi. 2021{\natexlab{a}}.
\newblock \href {https://doi.org/10.18653/v1/2021.naacl-main.46} {{XOR} {QA}: Cross-lingual open-retrieval question answering}.
\newblock In \emph{Proceedings of the 2021 Conference of the North American Chapter of the Association for Computational Linguistics: Human Language Technologies}, pages 547--564, Online. Association for Computational Linguistics.

\bibitem[{Asai et~al.(2021{\natexlab{b}})Asai, Yu, Kasai, and Hajishirzi}]{asai2021one}
Akari Asai, Xinyan Yu, Jungo Kasai, and Hanna Hajishirzi. 2021{\natexlab{b}}.
\newblock One question answering model for many languages with cross-lingual dense passage retrieval.
\newblock \emph{Advances in Neural Information Processing Systems}, 34:7547--7560.

\bibitem[{Asai et~al.(2021{\natexlab{c}})Asai, Yu, Kasai, and Hajishirzi}]{10.5555/3540261.3540839}
Akari Asai, Xinyan Yu, Jungo Kasai, and Hannaneh Hajishirzi. 2021{\natexlab{c}}.
\newblock One question answering model for many languages with cross-lingual dense passage retrieval.
\newblock In \emph{Proceedings of the 35th International Conference on Neural Information Processing Systems}, NIPS '21, Red Hook, NY, USA. Curran Associates Inc.

\bibitem[{Bonifacio et~al.(2022)Bonifacio, Jeronymo, Abonizio, Campiotti, Fadaee, Lotufo, and Nogueira}]{mmarco}
Luiz Bonifacio, Vitor Jeronymo, Hugo~Queiroz Abonizio, Israel Campiotti, Marzieh Fadaee, Roberto Lotufo, and Rodrigo Nogueira. 2022.
\newblock \href {https://arxiv.org/abs/2108.13897} {mmarco: A multilingual version of the ms marco passage ranking dataset}.
\newblock \emph{Preprint}, arXiv:2108.13897.

\bibitem[{Chen et~al.()Chen, Xiao, Zhang, Luo, Lian, and Liu}]{Chen_Xiao_Zhang_Luo_Lian_Liu}
Jianlv Chen, Shitao Xiao, Peitian Zhang, Kun Luo, Defu Lian, and Zheng Liu.
\newblock Bge m3-embedding: Multi-lingual, multi-functionality, multi-granularity text embeddings through self-knowledge distillation.

\bibitem[{Chirkova et~al.(2024)Chirkova, Rau, D{\'e}jean, Formal, Clinchant, and Nikoulina}]{chirkova-etal-2024-retrieval}
Nadezhda Chirkova, David Rau, Herv{\'e} D{\'e}jean, Thibault Formal, St{\'e}phane Clinchant, and Vassilina Nikoulina. 2024.
\newblock \href {https://doi.org/10.18653/v1/2024.knowllm-1.15} {Retrieval-augmented generation in multilingual settings}.
\newblock In \emph{Proceedings of the 1st Workshop on Towards Knowledgeable Language Models (KnowLLM 2024)}, pages 177--188, Bangkok, Thailand. Association for Computational Linguistics.

\bibitem[{DeepSeek-AI et~al.(2025)DeepSeek-AI, Guo, Yang, Zhang, Song, Zhang, Xu, Zhu, Ma, Wang, Bi, Zhang, Yu, Wu, Wu, and others.}]{deepseekai2025deepseekr1incentivizingreasoningcapability}
DeepSeek-AI, Daya Guo, Dejian Yang, Haowei Zhang, Junxiao Song, Ruoyu Zhang, Runxin Xu, Qihao Zhu, Shirong Ma, Peiyi Wang, Xiao Bi, Xiaokang Zhang, Xingkai Yu, Yu~Wu, Z.~F. Wu, and others. 2025.
\newblock \href {https://arxiv.org/abs/2501.12948} {Deepseek-r1: Incentivizing reasoning capability in llms via reinforcement learning}.
\newblock \emph{Preprint}, arXiv:2501.12948.

\bibitem[{Gao et~al.(2022)Gao, Yin, Li, Meng, Zhao, Yin, King, and Lyu}]{gao-etal-2022-retrieval}
Yifan Gao, Qingyu Yin, Zheng Li, Rui Meng, Tong Zhao, Bing Yin, Irwin King, and Michael Lyu. 2022.
\newblock \href {https://doi.org/10.18653/v1/2022.findings-naacl.92} {Retrieval-augmented multilingual keyphrase generation with retriever-generator iterative training}.
\newblock In \emph{Findings of the Association for Computational Linguistics: NAACL 2022}, pages 1233--1246, Seattle, United States. Association for Computational Linguistics.

\bibitem[{Grattafiori et~al.(2024)Grattafiori, Dubey, Jauhri, Pandey, Kadian, Al-Dahle, Letman, Mathur, Schelten, Vaughan, Yang, Fan et~al.}]{grattafiori2024llama3herdmodels}
Aaron Grattafiori, Abhimanyu Dubey, Abhinav Jauhri, Abhinav Pandey, Abhishek Kadian, Ahmad Al-Dahle, Aiesha Letman, Akhil Mathur, Alan Schelten, Alex Vaughan, Amy Yang, Angela Fan, and 1 others. 2024.
\newblock \href {https://arxiv.org/abs/2407.21783} {The llama 3 herd of models}.
\newblock \emph{Preprint}, arXiv:2407.21783.

\bibitem[{Kwiatkowski et~al.(2019)Kwiatkowski, Palomaki, Redfield, Collins, Parikh, Alberti, Epstein, Polosukhin, Devlin, Lee, Toutanova, Jones, Kelcey, Chang, Dai, Uszkoreit, Le, and Petrov}]{Kwiatkowski_NQ}
Tom Kwiatkowski, Jennimaria Palomaki, Olivia Redfield, Michael Collins, Ankur Parikh, Chris Alberti, Danielle Epstein, Illia Polosukhin, Jacob Devlin, Kenton Lee, Kristina Toutanova, Llion Jones, Matthew Kelcey, Ming-Wei Chang, Andrew~M. Dai, Jakob Uszkoreit, Quoc Le, and Slav Petrov. 2019.
\newblock \href {https://doi.org/10.1162/tacl_a_00276} {Natural questions: A benchmark for question answering research}.
\newblock \emph{Transactions of the Association for Computational Linguistics}, page 453–466.

\bibitem[{Lewis et~al.(2020)Lewis, Perez, Piktus, Petroni, Karpukhin, Goyal, Küttler, Lewis, Yih, Rocktäschel, Riedel, and Kiela}]{Lewis_2020}
Patrick Lewis, Ethan Perez, Aleksandara Piktus, Filippo Petroni, Vladimir Karpukhin, Naman Goyal, Heinrich Küttler, Mike Lewis, Wen-tau Yih, Tim Rocktäschel, Sebastian Riedel, and Douwe Kiela. 2020.
\newblock Retrieval-augmented generation for knowledge-intensive nlp tasks.
\newblock \emph{arXiv: Computation and Language,arXiv: Computation and Language}.

\bibitem[{Li et~al.(2024)Li, Haider, Luo, Agashe, and Callison-Burch}]{li-etal-2024-bordirlines}
Bryan Li, Samar Haider, Fiona Luo, Adwait Agashe, and Chris Callison-Burch. 2024.
\newblock \href {https://doi.org/10.18653/v1/2024.wikinlp-1.3} {{B}ord{IR}lines: A dataset for evaluating cross-lingual retrieval augmented generation}.
\newblock In \emph{Proceedings of the First Workshop on Advancing Natural Language Processing for Wikipedia}, pages 1--13, Miami, Florida, USA. Association for Computational Linguistics.

\bibitem[{Liu et~al.(2025)Liu, Trenous, Ribeiro, Byrne, and Hieber}]{liu-etal-2025-xrag}
Wei Liu, Sony Trenous, Leonardo F.~R. Ribeiro, Bill Byrne, and Felix Hieber. 2025.
\newblock \href {https://doi.org/10.18653/v1/2025.findings-emnlp.849} {{XRAG}: Cross-lingual retrieval-augmented generation}.
\newblock In \emph{Findings of the Association for Computational Linguistics: EMNLP 2025}, pages 15669--15690, Suzhou, China. Association for Computational Linguistics.

\bibitem[{Longpre et~al.(2021)Longpre, Lu, and Daiber}]{Longpre_Lu_Daiber_2021}
Shayne Longpre, Yi~Lu, and Joachim Daiber. 2021.
\newblock \href {https://doi.org/10.1162/tacl_a_00433} {Mkqa: A linguistically diverse benchmark for multilingual open domain question answering}.
\newblock \emph{Transactions of the Association for Computational Linguistics}, page 1389–1406.

\bibitem[{Loshchilov and Hutter(2019)}]{adamw}
Ilya Loshchilov and Frank Hutter. 2019.
\newblock \href {https://openreview.net/forum?id=Bkg6RiCqY7} {Decoupled weight decay regularization}.
\newblock In \emph{International Conference on Learning Representations}.

\bibitem[{Moon et~al.(2025)Moon, Kim, and Verma}]{moon-etal-2025-quality}
Hoyeon Moon, Byeolhee Kim, and Nikhil Verma. 2025.
\newblock \href {https://doi.org/10.18653/v1/2025.mrl-main.12} {Quality-aware translation tagging in multilingual {RAG} system}.
\newblock In \emph{Proceedings of the 5th Workshop on Multilingual Representation Learning (MRL 2025)}, pages 161--177, Suzhuo, China. Association for Computational Linguistics.

\bibitem[{Park and Lee(2025)}]{park-lee-2025-investigating}
Jeonghyun Park and Hwanhee Lee. 2025.
\newblock \href {https://doi.org/10.18653/v1/2025.findings-acl.295} {Investigating language preference of multilingual {RAG} systems}.
\newblock In \emph{Findings of the Association for Computational Linguistics: ACL 2025}, pages 5647--5675, Vienna, Austria. Association for Computational Linguistics.

\bibitem[{Qi et~al.(2025)Qi, Fern{\'a}ndez, and Bisazza}]{qi-etal-2025-consistency}
Jirui Qi, Raquel Fern{\'a}ndez, and Arianna Bisazza. 2025.
\newblock \href {https://doi.org/10.18653/v1/2025.mrl-main.15} {On the consistency of multilingual context utilization in retrieval-augmented generation}.
\newblock In \emph{Proceedings of the 5th Workshop on Multilingual Representation Learning (MRL 2025)}, pages 199--225, Suzhuo, China. Association for Computational Linguistics.

\bibitem[{Qwen et~al.(2025)Qwen, :, Yang, Yang, Zhang, Hui, Zheng, Yu, Li, Liu, Huang, Wei, Lin, Yang, Tu, Zhang, Yang, Yang, Zhou, Lin, Dang, Lu, Bao, Yang, Yu, Li, Xue, Zhang, Zhu, Men, Lin, Li, Tang, Xia, Ren, Ren, Fan, Su, Zhang, Wan, Liu, Cui, Zhang, and Qiu}]{qwen2025qwen25technicalreport}
Qwen, :, An~Yang, Baosong Yang, Beichen Zhang, Binyuan Hui, Bo~Zheng, Bowen Yu, Chengyuan Li, Dayiheng Liu, Fei Huang, Haoran Wei, Huan Lin, Jian Yang, Jianhong Tu, Jianwei Zhang, Jianxin Yang, Jiaxi Yang, Jingren Zhou, and 25 others. 2025.
\newblock \href {https://arxiv.org/abs/2412.15115} {Qwen2.5 technical report}.
\newblock \emph{Preprint}, arXiv:2412.15115.

\bibitem[{Ram et~al.(2023)Ram, Levine, Dalmedigos, Muhlgay, Shashua, Leyton-Brown, and Shoham}]{ram-etal-2023-context}
Ori Ram, Yoav Levine, Itay Dalmedigos, Dor Muhlgay, Amnon Shashua, Kevin Leyton-Brown, and Yoav Shoham. 2023.
\newblock \href {https://doi.org/10.1162/tacl_a_00605} {In-context retrieval-augmented language models}.
\newblock \emph{Transactions of the Association for Computational Linguistics}, 11:1316--1331.

\bibitem[{Yang et~al.(2024)Yang, J\"{a}nich, Mayfield, and Lawrie}]{peer}
Eugene Yang, Thomas J\"{a}nich, James Mayfield, and Dawn Lawrie. 2024.
\newblock \href {https://doi.org/10.1145/3626772.3657943} {Language fairness in multilingual information retrieval}.
\newblock In \emph{Proceedings of the 47th International ACM SIGIR Conference on Research and Development in Information Retrieval}, SIGIR '24, page 2487–2491, New York, NY, USA. Association for Computing Machinery.

\bibitem[{Zhang et~al.(2023)Zhang, Ogueji, Ma, and Lin}]{10.1145/3613447}
Xinyu Zhang, Kelechi Ogueji, Xueguang Ma, and Jimmy Lin. 2023.
\newblock \href {https://doi.org/10.1145/3613447} {Toward best practices for training multilingual dense retrieval models}.
\newblock \emph{ACM Trans. Inf. Syst.}, 42(2).

\bibitem[{Zhang et~al.(2025)Zhang, Li, Long, Zhang, Lin, Yang, Xie, Yang, Liu, Lin, Huang, and Zhou}]{qwen3embedding}
Yanzhao Zhang, Mingxin Li, Dingkun Long, Xin Zhang, Huan Lin, Baosong Yang, Pengjun Xie, An~Yang, Dayiheng Liu, Junyang Lin, Fei Huang, and Jingren Zhou. 2025.
\newblock Qwen3 embedding: Advancing text embedding and reranking through foundation models.
\newblock \emph{arXiv preprint arXiv:2506.05176}.

\bibitem[{Zhao et~al.(2025)Zhao, Huang, Hu, Wang, Mao, Zhang, Jiang, Wu, Ai, Wang, Zhou, and Chen}]{swift}
Yuze Zhao, Jintao Huang, Jinghan Hu, Xingjun Wang, Yunlin Mao, Daoze Zhang, Zeyinzi Jiang, Zhikai Wu, Baole Ai, Ang Wang, Wenmeng Zhou, and Yingda Chen. 2025.
\newblock \href {https://doi.org/10.1609/aaai.v39i28.35383} {Swift: A scalable lightweight infrastructure for fine-tuning}.
\newblock \emph{Proceedings of the AAAI Conference on Artificial Intelligence}, 39(28):29733--29735.

\end{thebibliography}

\appendix
\clearpage

\section{Metric Implementation Details}
\label{sec:metric_details}

We report some evaluation metrics in our experiments: character 3-gram Recall, Precision@k, NDCG@k, and PEER. Below, we describe their implementations in detail.

\paragraph{Character 3-gram Recall.}
Character 3-gram Recall measures the lexical coverage between the generated content and the reference text at the character level.
We extract all contiguous character 3-grams from both the reference text and the generated text.
Let $C_{\text{ref}}$ denote the multiset of character 3-grams from the reference, and $C_{\text{gen}}$ denote those from the generated text.
The character 3-gram Recall score is defined as:
\begin{equation}
\mathrm{Recall}_{\text{char-3}} =
\frac{| C_{\text{gen}} \cap C_{\text{ref}} |}{| C_{\text{ref}} |}
\end{equation}
This metric is robust to tokenization differences and is particularly suitable for multilingual evaluation.

\paragraph{Precision@k.}
Precision@k measures the proportion of relevant documents among the top-$k$ reranked results.
Formally, given a ranked list of documents $R_k$ of length $k$ and a binary relevance function $\mathrm{rel}(\cdot)$, Precision@k is defined as:
\begin{equation}
\mathrm{Precision@}k = \frac{1}{k} \sum_{i=1}^{k} \mathrm{rel}(R_i)
\end{equation}
where $\mathrm{rel}(R_i) = 1$ if the document at rank $i$ is relevant, and $0$ otherwise.

\paragraph{NDCG@k.}
Normalized Discounted Cumulative Gain (NDCG@k) takes into both the relevance and the ranking position of documents.
We first compute DCG@k as:
\begin{equation}
\mathrm{DCG@}k = \sum_{i=1}^{k} \frac{2^{\mathrm{rel}(R_i)} - 1}{\log_2(i + 1)}
\end{equation}
where $\mathrm{rel}(R_i)$ denotes the relevance score of the document at rank $i$.
NDCG@k is obtained by normalizing DCG@k with the ideal DCG@k (IDCG@k), which corresponds to the optimal ranking:
\begin{equation}
\mathrm{NDCG@}k = \frac{\mathrm{DCG@}k}{\mathrm{IDCG@}k}
\end{equation}
This normalization ensures that NDCG@k ranges between $0$ and $1$.

\begin{figure}[t]
    \centering
    \includegraphics[width=\linewidth]{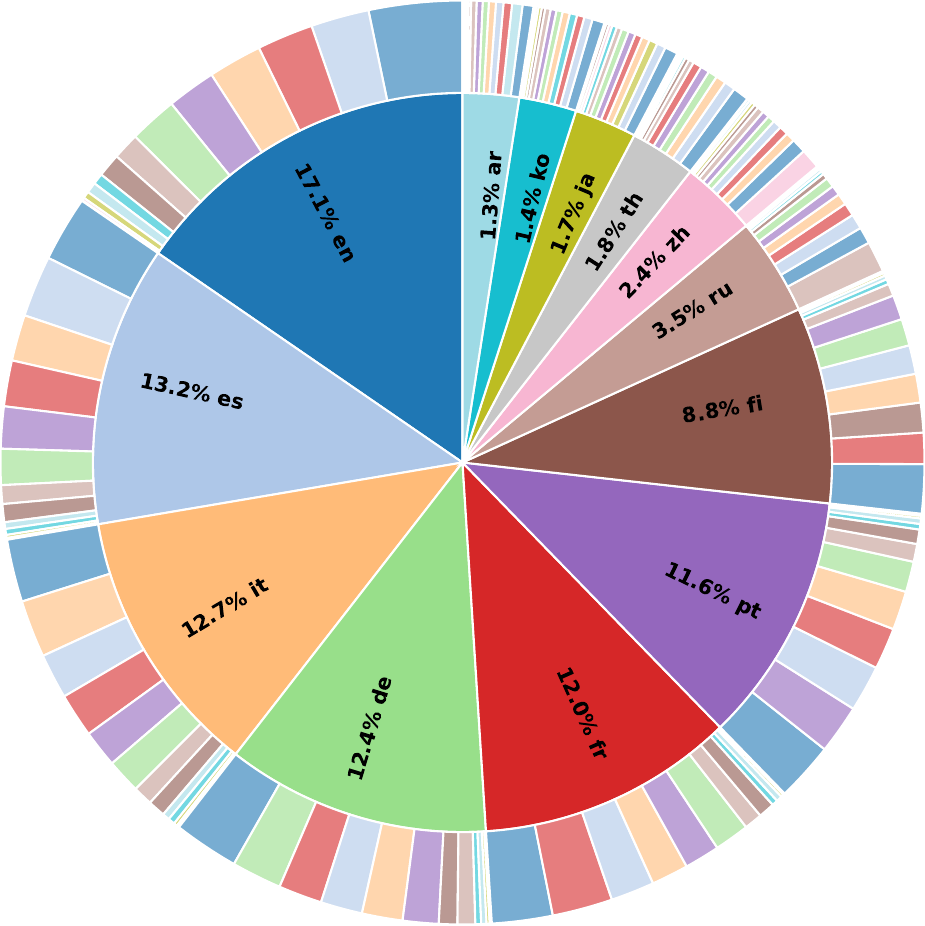}
    \caption{Language distribution of queries (inner ring) and positive documents (outer ring).}
    \label{fig:lan_dis_of_dataset}
\end{figure}

\paragraph{PEER.}
We compute PEER (Probability of Equal Expected Rank) following \citet{peer}, with a task-specific adaptation: we only use positive documents in the fairness test.
Intuitively, PEER evaluates whether relevant documents written in different languages receive systematically different ranks.

For each query $q$, we collect all retrieved documents labeled as positive and record their rank positions in the final ranked list.
We then partition these ranks by the document language $\ell \in \mathcal{L}$, yielding groups $\{ \mathcal{R}_{q,\ell} \}_{\ell \in \mathcal{L}}$, where $\mathcal{R}_{q,\ell}$ is the multiset of rank positions of positive documents in language $\ell$.

We apply the Kruskal--Wallis $H$ test (KW) on these rank groups, with the null hypothesis that the rank distributions of positive documents are identical across languages (i.e., equal expected ranks).
We define PEER for query $q$ as the resulting $p$-value:
\begin{equation}
\mathrm{PEER}(q) = p\big(\mathrm{KW}(\{\mathcal{R}_{q,\ell}\}_{\ell \in \mathcal{L}})\big)
\end{equation}
where higher values (closer to $1$) indicate that we cannot reject the hypothesis of equal expected rank, suggesting better language fairness. We report the final PEER score as the mean of $\mathrm{PEER}(q)$ over all queries.

\section{Statistics of the LAURA Dataset}
\label{sec:stat-of-finetuning-dataset}

\begin{table}[t]
    \centering
    \resizebox{0.48\textwidth}{!}{
    \begin{tabular}{lcc}
    \toprule
    &Train Set & Dev Set\\
    \midrule
    Queries&17,360 &1,000 \\
    Positive Documents & 114,867& 6,762\\
    Avg. Languages per Query &2.90 & 2.90\\
    \bottomrule
    \end{tabular}
    }
    \caption{Statistics of the LAURA dataset. Avg. Languages per Query indicates the average number of distinct languages among the positive candidate documents associated with each query.}
    \label{tab:stat_of_dataset}
\end{table}

\begin{table}[t]
    \centering
    \resizebox{0.48\textwidth}{!}{
    \begin{tabular}{lccccc}
        \toprule
        Language & Baseline & LAURA & Δ\% & p \\ 
        \midrule
        Portuguese & 63.1 & 65.9 & +4.4 & 1.24e-24*** \\ 
        Finnish & 55.1 & 57.8 & +4.7 & 1.15e-17*** \\ 
        French & 59.8 & 62.3 & +4.3 & 1.26e-19*** \\ 
        Spanish & 62.3 & 64.8 & +4.0 & 2.94e-20*** \\ 
        Italian & 61.1 & 63.5 & +4.1 & 1.70e-18*** \\ 
        German & 60.3 & 62.1 & +2.9 & 1.46e-10*** \\ 
        Arabic & 31.7 & 32.9 & +3.8 & 5.22e-04*** \\ 
        English & 66.6 & 67.8 & +1.7 & 8.68e-06*** \\ 
        Russian & 46.8 & 47.5 & +1.5 & 4.62e-02* \\ 
        Thai & 24.2 & 24.9 & +2.9 & 1.20e-02* \\ 
        Japanese & 28.1 & 28.6 & +1.8 & 1.02e-01 \\ 
        Chinese & 25.3 & 25.8 & +1.8 & 8.04e-02 \\ 
        Korean & 25.0 & 24.8 & -0.8 & 4.98e-01 \\ 
        Overall & 46.9 & 48.4 & +3.1 & 8.62e-74*** \\ 
        \bottomrule
    \end{tabular}
    }
    \caption{Per-language paired $t$-test results on 3-gram scores.}
    \label{tab:significance}
\end{table}

\begin{table}[t]
    \centering
    \small
    \resizebox{0.48\textwidth}{!}{
    \begin{tabular}{lcccc}
    \toprule
    \multirow{2}{*}[-0.8ex]{\textbf{Setting}} & \multicolumn{2}{c}{\textbf{Llama}} & \multicolumn{2}{c}{\textbf{Qwen}} \\
    \cmidrule(lr){2-3} \cmidrule(lr){4-5}
     & \textbf{3-gram} & \textbf{Pearson}& \textbf{3-gram} & \textbf{Pearson} \\
    \midrule
        BGE-Reranker  &  48.9 &0.198 & 46.7 &0.188 \\
        Stage 1  &  48.7 &\textbf{0.272} & 46.5 &\textbf{0.269} \\
        LAURA & \textbf{49.9} &0.236 & \textbf{47.7} & 0.247\\ 
    \bottomrule
    \end{tabular}
    }
    \caption{Ablation study comparing the full pipeline against using only Stage 1 training data.}
    \label{tab:ablation_stages}
\end{table}

As shown in Table~\ref{tab:stat_of_dataset}, we report the number of queries and positive documents in the constructed LAURA dataset, as well as the average number of languages per query among the candidate positive documents. Figure~\ref{fig:lan_dis_of_dataset} illustrates the language distributions of queries and positive documents.

\section{Detailed MKQA test Results}
To facilitate comparison, Table~\ref{tab:mkqa_xor_upper} reports detailed per-language results on the MKQA test set used in the main experiments.
\begin{table*}[t]
\centering
% \small
\resizebox{\textwidth}{!}{
\begin{tabular}{l|ccccccccccccc|c}
\toprule
\textbf{Setting}
& ar & de & en & es & fi & fr & it & ja & ko & pt & ru & th & zh & \textbf{Avg.} \\

\midrule
\multicolumn{15}{c}{\textbf{\textit{Llama3-8B-Instruct}}} \\
\midrule

Upper-limit& 53.6 & 76.5 & 79.3 & 76.8 & 73.4 & 76.7 & 77.0 & 47.9 & 41.0 & 78.4 & 68.0 & 44.1 & 33.8 & 63.6 \\
BGE-Reranker   & 32.7  & 62.8  & 70.1  & 63.0  & 58.1  & 64.4  & 63.9  & 29.2  & 25.5  & 66.4  & 51.9  & 26.4  & 21.7  & 48.9   \\
\quad\textbf{+ LAURA} & 33.2  & 64.8  & 69.8  & 65.7  & 59.9  & 66.7  & 65.5  & 28.9  & 24.9  & 69.1  & 50.8  & 26.9  & 22.2  & \textbf{49.9}  \\
Qwen-Reranker& 28.9 & 60.9  & 67.8 & 62.7 & 54.2 & 63.7 & 62.2 & 28.2 & 23.3 & 66.3 & 47.4 & 24.8 & 21.9 & 47.1 \\ 
\quad\textbf{+ LAURA} & 31.6  & 63.3  & 70.1  & 65.5  & 57.3  & 66.1  & 65.0  & 28.2  & 23.8  & 68.4  & 50.5  & 26.7  & 22.7  & 49.2  \\ 
\midrule
\multicolumn{15}{c}{\textbf{\textit{Qwen2.5-7B-Instruct}}} \\
\midrule

Upper-limit & 51.4 & 73.7 & 76.2 & 75.6 & 71.5 & 71.6 & 73.8 & 44.9 & 38.9 & 73.8 & 63.1 & 39.0 & 42.8 & 61.3 \\
BGE-Reranker   & 33.8  & 59.6  & 65.4  & 62.3  & 55.5  & 56.6  & 60.2  & 28.0  & 26.5  & 60.5  & 45.7  & 23.5  & 29.0  & 46.7  \\ 
\quad\textbf{+ LAURA} & 33.9  & 61.0  & 66.0  & 64.2  & 58.4  & 58.8  & 61.8  & 29.4  & 25.7  & 63.4  & 44.8  & 23.4  & 29.5  & \textbf{47.7} \\
Qwen-Reranker & 31.4 & 58.0  & 63.3 & 61.3 & 52.8 & 54.4 & 57.9 & 27.2 & 24.7 & 59.3 & 42.3 & 22.3 & 28.7 & 44.9\\ 
\quad\textbf{+ LAURA} & 33.1  & 59.2  & 65.2  & 63.9  & 55.3  & 57.8  & 61.9  & 28.1  & 25.0  & 62.6  & 44.1  & 22.7  & 28.7  & 46.7 \\ 
\bottomrule
\end{tabular}
}
\vspace{-3pt}
\caption{Performance comparison between the vanilla document reranking and the oracle evidence estimation settings on MKQA.
All results are reported using character 3-gram recall. \textbf{Bolded results} denote the best performance among all non–upper-limit settings.}
\label{tab:mkqa_xor_upper}
\end{table*}

\section{Case Study}
Table~\ref{tab:case_study} presents a case study illustrating the limitations of relevance-based reranking in the vanilla multilingual RAG setting.
\label{sec: case-study}
\begin{table*}[t]
\centering
\small
\renewcommand{\arraystretch}{1.4}
\begin{tabularx}{\textwidth}{l|X}
\toprule
\textbf{Query} & \textbf{Who plays the blue lady in The Fifth Element?} \\
\midrule
\textbf{Label} & Maïwenn \\
\midrule
\midrule

\textbf{Vanilla Top-5} & 
\textbf{[1] (es)} El quinto elemento. El quinto elemento (en francés: Le Cinquième Élément) es una película francesa (con coproducción de EE.UU.) de ciencia ficción y acción de 1997 dirigida por Luc Besson, con \textbf{Bruce Willis, Milla Jovovich y Gary Oldman en los papeles principales.} Principalmente ambientada en el , la trama central de la película involucra la supervivencia del planeta Tierra, que se convierte en responsabilidad de Korben Dallas (Willis)... \newline
\textbf{[2] (en)} The Fifth Element. The Fifth Element The Fifth Element () is a 1997 French science fiction action film directed and co-written by Luc Besson. \textbf{It stars Bruce Willis, Gary Oldman and Milla Jovovich. Primarily set in the 23rd century}, the film's central plot involves the survival of planet Earth, which becomes the responsibility of Korben Dallas (Willis), a taxicab driver and former special forces major, \textbf{after a young woman (Jovovich) falls into his cab}... \newline 
\textbf{[3] (it)} Il quinto elemento. Il quinto elemento (Le Cinquième Élément) è un film del 1997 diretto da Luc Besson. Di produzione francese (benché girato in lingua inglese), fu la pellicola più costosa mai prodotta in Europa all\'epoca della sua uscita. \textbf{Il film, che ha per protagonisti Bruce Willis, Milla Jovovich e Gary Oldman}, venne presentato fuori concorso al 50º Festival di Cannes... \newline 
\textbf{[4] (ru)} {\rufont Пятый элемент (фильм). того, для создания костюмов был привлечён известный модельер Жан-Поль Готье. Он разработал все 900 костюмов, использованных в сценах на корабле «Флостон Парадайз». \textbf{Костюм Лилу из белых полос ткани Готье создал}, вдохновившись картиной Фриды Кало «Сломанная колонна». В течение года команда создала более 8000 рисунков. В это время Бессон предложил на главную роль Брюса Уиллиса и Мела Гибсона, а также рассматривал \textbf{Джулию Робертс на роль Лилу}... }\newline
\textbf{[5] (zh)} 第五元素(電影). \textbf{米拉·乔沃维奇饰）的人形女性。莉露对周围的一切深感恐惧，逃出实验室后，她从楼层的外沿跳了下去}，正好掉进前特种部队少校科本·达拉斯（布鲁斯·威利斯饰）所开的出租车裡... \\
\midrule
\textbf{Model Answer} & Milla Jovovich plays the blue lady, Leeloo, in The Fifth Element. \textcolor{red}{Wrong} \\
\midrule
\midrule

\textbf{Oracle Top-5} &
\textbf{[1] (de)} Das fünfte Element. Das fünfte Element (Originaltitel: Le Cinquième Élément) ist ein Science-Fiction-Film von Luc Besson mit Bruce Willis und Milla Jovovich aus dem Jahr 1997. Das fünfte Element ist aufgrund seiner hohen Einspielergebnisse von über 260 Millionen US-Dollar einer der bisher kommerziell erfolgreichsten europäischen Filme. Handlung Der Film beginnt im Jahr 1914 in Ägypten, in einem verfallenen Tempel, wo der Archäologe Professor Pacoli, begleitet vom Reporter Billy und einem Priester, Inschriften über das unfassbar Böse findet... \newline 
\textbf{[2*] (de) (rank 10 in baseline)} Das fünfte Element. der Antagonist Zorg begegnen sich im Film kein einziges Mal. Die Kostüme und Accessoires wurden von dem französischen Modeschöpfer Jean Paul Gaultier entworfen. Als sich der Archäologe zu Beginn des Films plötzlich riesigen Mondoshawan-Aliens gegenübersieht, fragt er in der deutschen Fassung: „Sind Sie hier von der Erde?“, während es im Original heißt: „Are you German?“ (dt. „Sind Sie Deutsche(r)?“). Der erste Teil der Arie der Diva ist aus der Oper Lucia di Lammermoor von Gaetano Donizetti und wird hier von Inva Mula gesungen. \textbf{Als Darstellerin der Diva agierte jedoch Maïwenn}, mit der Regisseur Besson zum Zeitpunkt der Dreharbeiten zusammenlebte und...\newline
\textbf{[3] (de)} Milla Jovovich. Milica „Milla“ Jovovich (* 17. Dezember 1975 in Kiew, Ukrainische SSR, Sowjetunion, ukrainisch {\rufont Милиця Богданівна Йовович}) ist eine US-amerikanische Schauspielerin und Model serbisch-russischer Herkunft. Bekannt wurde sie nach Erfolgen in den Filmen Das fünfte Element und Johanna von Orleans, aber besonders für ihre Hauptrolle in der Filmreihe Resident Evil... \newline 
\textbf{[4] (de)} Das fünfte Element. ins Weltall geschossen wird. In einer abschließenden Szene will sich der Präsident bei den beiden „Helden“ bedanken, die sich aber leidenschaftlich lieben und deshalb unabkömmlich sind. Auszeichnungen Der Film wurde im Jahr 1998 für den Oscar in der Kategorie Bester Tonschnitt nominiert. Er wurde 1998 in den Kategorien Bester Science-Fiction-Film, Beste Spezialeffekte, Beste Kostüme und Beste Nebendarstellerin (Milla Jovovich) für den Saturn Award nominiert... \newline 
\textbf{[5] (de)} Das fünfte Element. den Kategorien Bester Film, Beste Kostüme, Bester Schnitt, Beste Filmmusik und Bester Ton für den gleichen Preis nominiert. Milla Jovovich wurde 1998 für den Blockbuster Entertainment Award und (für die Beste Kampfszene) den MTV Movie Award nominiert. Der Film gewann 1997 die Goldene Leinwand, den Bogey Award in Silber und wurde für den Europäischen Filmpreis nominiert...\\

\midrule
\textbf{Model Answer} & The role of the blue alien diva Plavalaguna was played by Maïwenn. \textcolor{green}{True}\\

\bottomrule
\end{tabularx}
\caption{A case study revealing a limitation of relevance-based reranking in multilingual RAG. The answer-critical document (marked as \textbf{[2*]}) is retrieved but ranked only 10th under the baseline, causing relevance-based reranking to produce an incorrect answer.}
\label{tab:case_study}
\end{table*}

\section{Language Distribution and Model Performance}
\label{sec: all-lang-dist-fig}
\begin{figure*}[t]
  \centering
  \includegraphics[width=\textwidth]{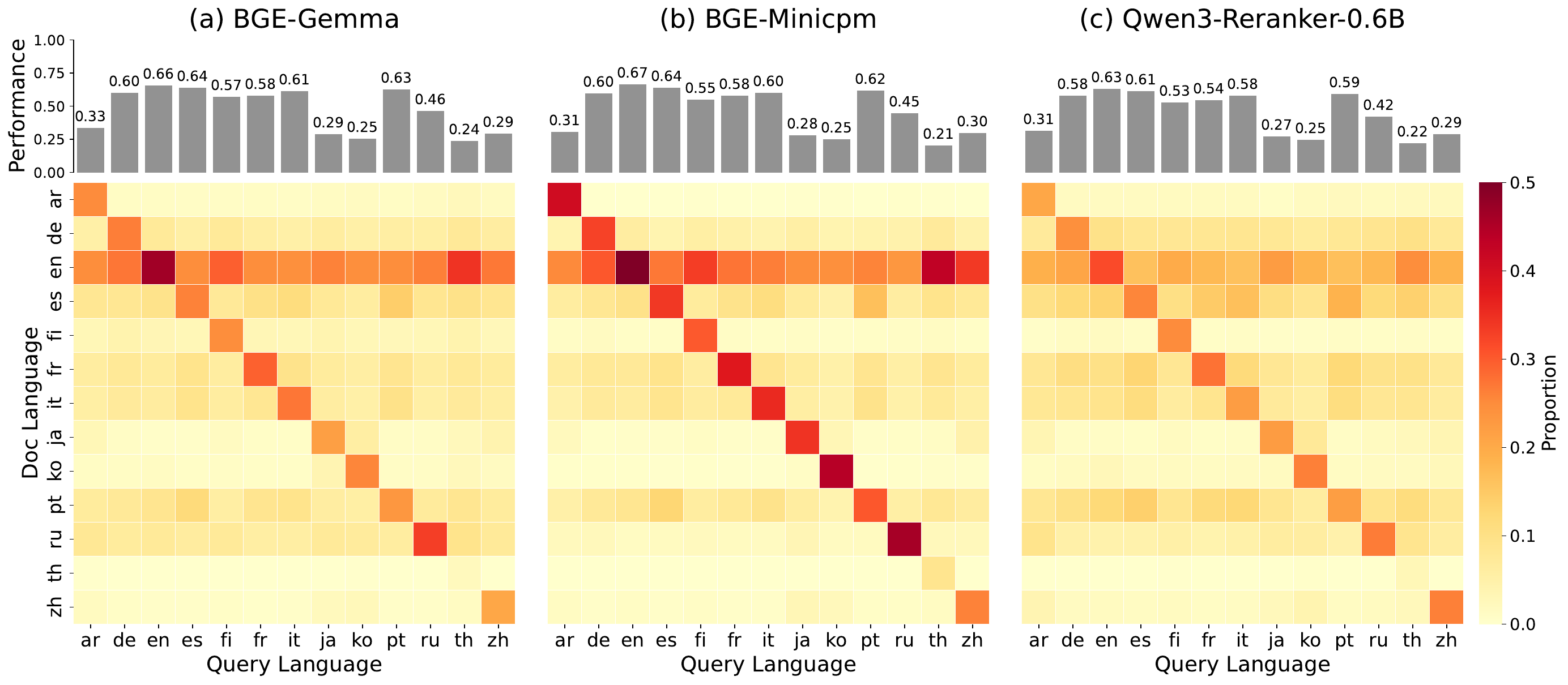}
  \caption{Vanilla document reranking with BGE-emma, BGE-Minicpm and Qwen3-Reranker-0.6B rerankers. The heatmap shows the language distribution, while the bar chart reports Recall@3-gram of Qwen2.5-7B-Instruct.
  }
  \label{fig:all-venilla}
  % \vspace{-12pt}
\end{figure*}

\begin{figure*}[t]
  \centering
  \includegraphics[width=\textwidth]{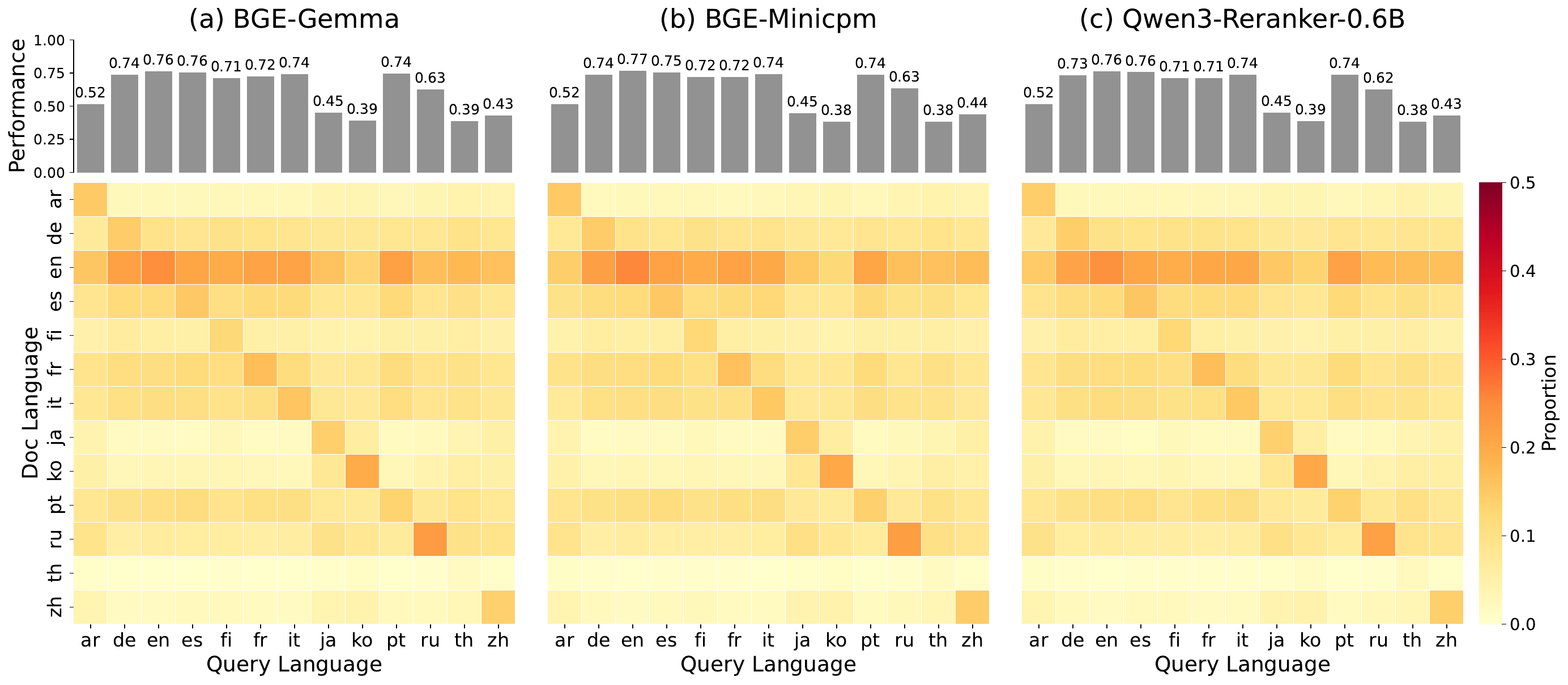}
  \caption{Oracle evidence estimation with BGE-Gemma, BGE-Minicpm and Qwen3-Reranker-0.6B rerankers. The heatmap shows the language distribution, while the bar chart reports Recall@3-gram of Qwen2.5-7B-Instruct.
  }
  \label{fig:all-upper}
  % \vspace{-12pt}
\end{figure*}

Figure~\ref{fig:all-venilla} and Figure~\ref{fig:all-upper} show the document language distribution and generation performance under the vanilla and upper-limit settings, respectively, across different rerankers.

\section{Statistical Significance of LAURA Improvements}
To assess whether the RAG performance gains from LAURA are statistically reliable, we compute per-query 3-gram scores and construct paired samples between LAURA and the corresponding baseline under identical configurations (2 rerankers $\times$ 2 generators), yielding approximately 4,000 paired observations per language. We apply two-tailed paired $t$-tests on the per-query score differences. Results are reported in Table~\ref{tab:significance}. LAURA achieves statistically significant gains ($p < 0.05$) in 10 out of 13 languages, and the overall improvement is highly significant ($p = 8.62 \times 10^{-74}$). All conclusions hold under Bonferroni correction, confirming that the improvements reflect a systematic effect rather than sampling variation.

\section{Ablation of LAURA}
To further validate the necessity of both Stage 1 and Stage 2 in our pipeline, we conduct an ablation study using the training data obtained from Stage 1 alone. 

As shown in Table~\ref{tab:ablation_stages}, training solely with the data from Stage 1 improves the correlation coefficient. However, due to the lack of filtering, a substantial number of false-positive documents are included, making it difficult to achieve meaningful improvements in downstream performance. In contrast, Stage 2 performs document-level evaluation and filtering, which substantially improves data quality and consequently leads to further gains in the final generation performance.

\end{document}